# REFLECTIVE HYBRID INTELLIGENCE FOR MEANINGFUL HUMAN CONTROL IN DECISION-SUPPORT SYSTEMS

**Authors:** Catholijn M. Jonker, Luciano Cavalcante Siebert and Pradeep K. Murukannaiah

**Author ORCIDs:** 0000-0003-4780-7461, 0000-0002-7531-3154, 0000-0002-1261-6908

**Abstract:** With the growing capabilities and pervasiveness of AI systems, societies must collectively choose between reduced human autonomy, endangered democracies and limited human rights, and AI that is aligned to human and social values, nurturing collaboration, resilience, knowledge and ethical behaviour. In this chapter, we introduce the notion of self-reflective AI systems for meaningful human control over AI systems. Focusing on decision support systems, we propose a framework that integrates knowledge from psychology and philosophy with formal reasoning methods and machine learning approaches to create AI systems responsive to human values and social norms. We also propose a possible research approach to design and develop self-reflective capability in AI systems. Finally, we argue that self-reflective AI systems can lead to self-reflective hybrid systems (human + AI), thus increasing meaningful human control and empowering human moral reasoning by providing comprehensible information and insights on possible human moral blind spots.

**Keywords:** Automated negotiation, Fairness, Hybrid intelligence, Moral Reasoning, Reflective equilibrium, Self-adaptive systems.

## 11.1 Introduction

The narrative that Artificial Intelligence (AI) systems are (or soon will be) able to outperform and replace humans in most tasks is increasing in the popular media (e.g., a language generator being able to write a coherent essay (GPT-3, 2020), self-driving cars enhancing road safety (Marshall, 2017)). However, such narratives disregard the limitations of AI systems in properly identifying, estimating and aligning to inherent human concepts such as moral values, social norms, emotions and creativity. Persisting on this road of replacing humans by machines can lead to a future with severe violation to basic human rights, endangered democracies, increased political polarization, manipulation and disinformation.

Central to this discussion is the concept of autonomy. From the perspective of moral and political philosophy, autonomy puts weight on a person's ability to self-govern (Christman, 2020). From a technical perspective, autonomy relates to the capacity of an artificial agent to operate independently of human guidance (Totschnig, 2020). In contrast, we view autonomy from a hybrid (human and AI) systems perspective, in which human intellect is augmented (not replaced) by artificial agents (Akata et al., 2020). To achieve this, meaningful human control is crucial (Article 36, 2014; 2015; Horowitz & Scharre, 2015; Santoni de Sio & van den Hoven, 2018).

An important question, then, is: how can humans remain in control of AI-based systems designed to operate autonomously (Siebert et al., 2018)? In line with Bradshaw's idea about "the seven deadly myths of autonomous systems" (Bradshaw et al., 2013; Johnson et al., 2014b; Hoffman, Hawley, & Bradshaw, 2014), we argue that the concept of autonomy must be considered from a socio-technical perspective—looking into not only the functional but also cognitive, social and moral (in)abilities of human and AI agents—to achieve control over an AI system. That is, AI agents should be designed to operate as humans' interdependent teammates (Johnson et al., 2014a), so that human control can be meaningful.

---

[1] Jonker, C.,M. Siebert, L. C., Murukannaiah, P. K. Reflective hybrid intelligence for meaningful human control in decision support systems (Forthcoming 2023). Research Handbook on Meaningful Human Control of Artificial Intelligence Systems. Publisher: Edward Elgar Publishing.

Considering the philosophical account on meaningful human control proposed by Santoni de Sio and van den Hoven (2018), we can analyze the interdependence between human and AI agents with respect to the *tracking*[1] and *tracing*[2] conditions. To track moral reasons, AI agents need to monitor and adapt their behaviour to new (social) circumstances in dynamic environments. However, as tracing a moral reason requires human moral understanding of the system's behaviour, agents should not perform such tasks independently, i.e., without human collaboration, supervision or at least awareness. If an agent adapts itself "autonomously", relevant humans would likely be unable to maintain a moral understanding of the system.

Further, unexpected and morally challenging situations might arise, where we do not want machines to make a decision at all. In these situations, humans should be expected to engage in proper moral reflection to reach a morally acceptable solution. Looking at the implications of the tracking and tracing conditions in hybrid systems, it becomes clear that some form of monitoring and continuous improvement is needed to align humans and AI agents. Thus, we need some form of Reflective Hybrid Intelligence (RHI).

The topic of reflection has been discussed from many perspectives including computer science (Steunebrink & Schmidhuber, 2012; Tomforde et al., 2014) and philosophy. From a computer science perspective, self-reflection is usually defined as ability of the system to continuously monitor and improve its own behaviour (Tomforde et al., 2014). From an ethical perspective, the goal of (self-) reflection is to come to a well-argued choice among various options for actions (van de Poel & Royakkers, 2007). In this work, we consider hybrid (human + AI) reflection towards a set of moral values as understood by a given person or a group of people. The part of the self-reflection that the agent has to do requires a fully computational model. For this, we consider the MAPE-K (Monitor-Analyze-Plan-Execute over a shared Knowledge) feedback loop (Kephart & Chess, 2003) as a starting point. For RHI, human stakeholders must also perform self-reflection and be aware of the decision-making processes of the agent to achieve meaningful human control. To account for moral reasoning processes which cannot (and, arguably, should not) be automated fully, we consider the elements of Wide Reflective Equilibrium (WRE) (Rawls, 2003; Welch, 2014), the contemporary orthodoxy for reflection on applied ethics, as the "K" (Knowledge base) in the MAPE-K loop. We argue that RHI cannot only increase meaningful human control but also empower human moral reasoning by providing comprehensible information and insights on possible human moral blind spots. We discuss and evaluate our framework for hybrid self-reflective systems in the context of decision support systems. In particular, we consider of humans negotiating with each other using a negotiation support system (Jonker et al., 2017), in order to achieve fair outcomes. The remainder of this chapter is structured as follows. Section 11.2 presents the theoretical background of MAPE-K and WRE. Section 11.3 presents our approach for Reflective Hybrid Intelligent (RHI) systems to support meaningful human control. Section 11.4 presents a detailed working example in the context of negotiation support systems. Section 11.5 concludes the chapter.

**11.2 Background**

In this section, we present and discuss reflective processes, both from an agent (11.2.1) and a human perspective (11.2.2) with a focus on moral aspects of reflection.

*11.2.1 Agent reflection processes*
Tomforde et al. (2014) define computational self-reflection as a computational system's ability "to continuously monitor and improve its own behaviour in an uncertain, dynamic, and time-invariant environment for situations that may not have been anticipated at design-time of the system." Further, they specify three requirements for a system to be called self-reflective. These include: the system's ability to (1) monitor the environment and the system's own behaviour, (2) model the system's knowledge (and metaknowledge) about the environment and behaviour, and (3) define new goals and ways to achieve those.

We adopt the above formulation of self-reflection, but with a main difference. In Tomforde et al.'s formulation, humans may be part of the self-reflection process, e.g., as a knowledge source. In contrast, in our formulation humans are an essential part of the self-reflection process. That is, we see each of the three requirements for self-reflection as interdependence requirements (Johnson et al, 2014a) between humans and computational entities (such as AI agents).

In order to facilitate self-reflection, we start with a model of the system based on MAPE-K (Kephart & Chess, 2003), a well-known reference control model used in auto nomic computing and self-adaptive systems (Arcaini, Riccobene, & Scandurra, 2015). In the MAPE-K model, a system consists of *autonomic elements* that interact with each other. Each autonomic element consists of a manager and one or more managed elements (which can be hardware or software resources). The manager of the autonomic element is modelled as a feedback loop consisting of monitor, analyze, plan and execute modules. Thus, an autonomic entity monitors its managed entity and the surrounding environment, analyzes the monitored information and other knowledge, and constructs and executes plans based on the analyses.

The original vision of the MAPE-K model was to relieve humans of the responsibility of directly managing the managed elements. Although it is useful to relieve humans from this responsibility for certain elements, it is necessary to keep humans responsible for certain elements, e.g., where managing an element has moral implications. Thus, we seek to adapt the MAPE-K model to keep humans in meaningful control of a managed element, where necessary.

*11.2.2 Moral reasoning and reflective processes*
Moral reasoning is a kind of practical reasoning, where one should decide, through reflection, what one ought to do morally. It involves questions about what is right or wrong, and virtuous or vicious. The term reflective equilibrium was coined in (Rawls, 2004) by discussing the problem of distributive justice. In a situation, there can be a broad set of judgements held by the people involved, which taken together may conflict with acceptable principles, rules or convictions. Going back and forth, and altering the situation, judgements or principles, we may find a situation that expresses reasonable conditions and yields principles which match our judgements (Rawls, 2004, 2013; Miller, 2021). A set of relevant background theories (scientific and philosophical) is needed to provide independent support to moral judgements and principles, widening the reflective equilibrium. By going back and forth between these three aspects and making adjustments, one arrives at an equilibrium (Figure 11.1). WRE may assist in producing greater moral agreement by bringing together ethical theory with practical ethics and, through the background theories, render problems more tractable (Daniels, 1996).

The WRE method may serve different purposes and be carried out by individuals acting together or separately. Collective approaches are widely applied for agreement among stakeholders in morally loaded situations, especially on medical ethics (Daniels, 1996), where knowledge of the specific situation must be included in the equilibrium. In situations where individuals work separately, for example, in thinking about the course of right action in a particular case, WRE can support the alignment of reasons and principles that one may appeal to and are notoriously general but lack context-dependency with particular moral judgements and theories. However, WRE neither provides, nor pretends to, quasi-algorithmic procedures for moral decision making. Its relevance lies in the regulative ideal of communicative transparency of justification (van den Hoven, 1997).

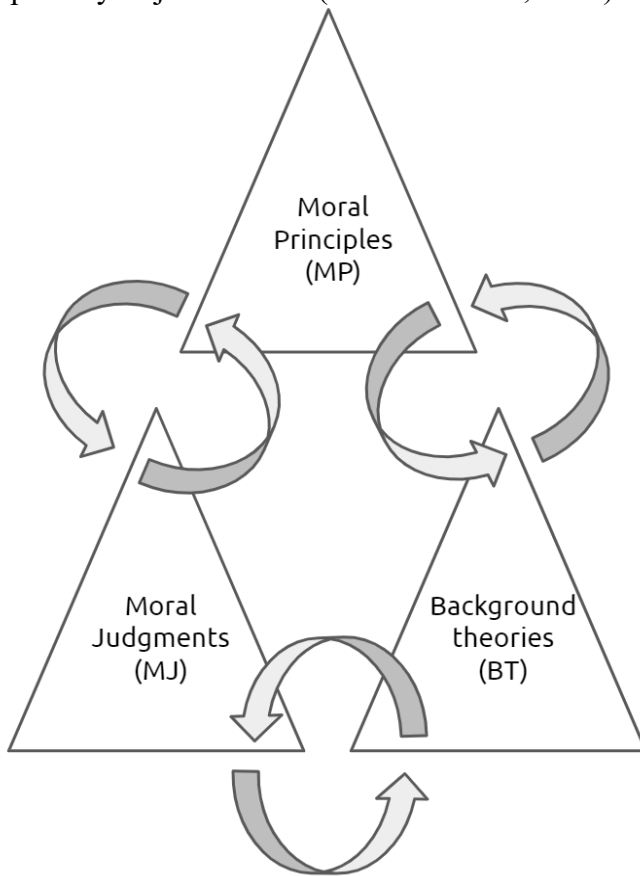

*Figure 11.1* Method of Wide Reflective Equilibrium (WRE)

The method of reflective equilibrium has inspired developments on responsible and ethical AI. Yudkowsky (2004) describes the concept of coherent extrapolated volition, where an agent tries to extrapolate human morality under a process of idealization. Yilmaz et al. (2016) approach reflective equilibrium via Thagard's Coherence theory to manage multiple principles in a context-sensitive manner in order to support both the development of an artificial ethical advisor and for designing decision-making models. Instead of approaching reflective equilibrium as means to design internal moral reasoning capacities of an agent, Burton et al. (2020) consider it from a design perspective to navigate issues arising from the semantic gap—particularly considering manufacturers' responsibility for harms. In this work, we take an approach that aims to combine reflection abilities of AI agents, e.g., through the MAPE-K framework, with the advantages of the reflective equilibrium method.

**11.3 Reflective Hybrid Intelligent (RHI) systems**
We focus on hybrid reflection on values. Reflection on values as opposed to functional aspects of a task is challenging due to the inherent ambiguity of what a value means for someone in a given context. In other words, values are often considered as pluralistic concepts (Chang, 2015), which may be incommensurable (Hsieh & Andersson, 2021). Based on the MAPE-K architecture (Kephart & Chess, 2003) and the method of Wide Reflective Equilibrium (WRE) (Rawls, 2003; Welch, 2014), we propose that humans and machines should jointly engage on reflection through the MAPE-K loop in order to achieve morally desirable outcomes from the human perspective (Figure 11.2).

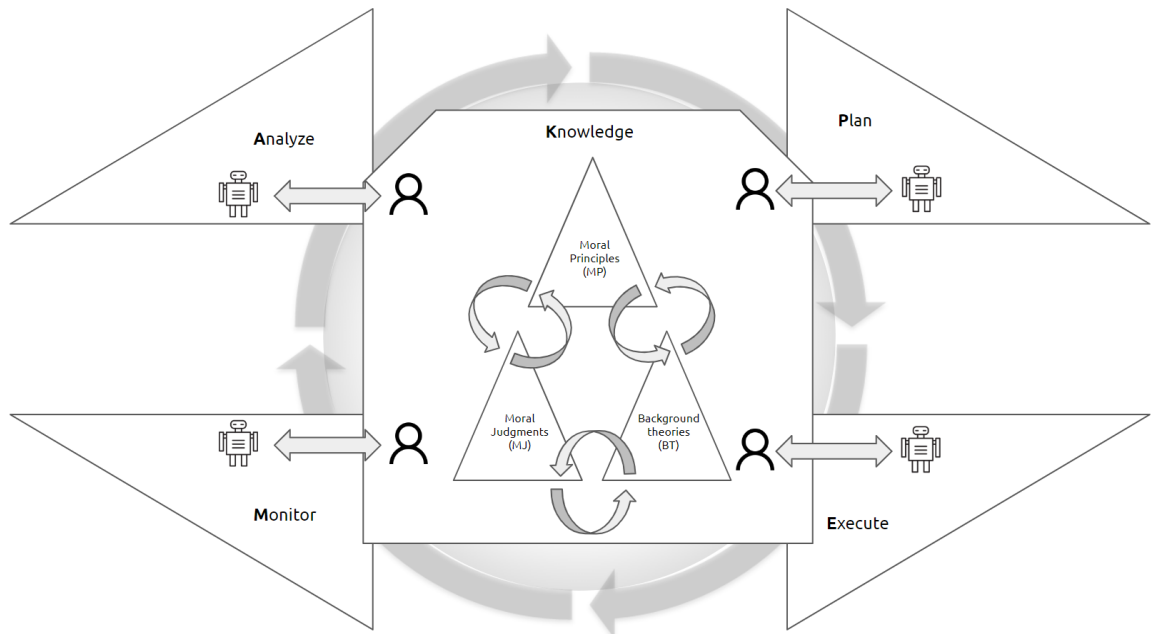

*Figure 11.2* Reflective Hybrid Intelligent (RHI) system based on the MAPE-K architecture and the method of Wide Reflective Equilibrium (WRE)

Humans trying to independently perform reflection on a socio-technical system (Murukannaiah, 2020) can be challenging due to cognitive overload and lack of situational awareness. Consider, for example, a fraud detection algorithm that 5nalyses thousands of records every second, or an automated vehicle that requests the driver to take control abruptly. Further, if a hybrid system rarely encounters situations which require reflection on moral values, the humans might get distracted or not motivated to engage on the reflection processes (van der Waa et al., 2020). However, attributing the responsibility of reflection (on values) to AI agents is even more problematic, as these agents might not have the nuanced and context dependent ability to interpret and reason about moral values. It seems that one is then caught between a rock and a hard place, where meaningful human control seems not achievable. However, hybrid reflection, in which the human and the agent complement each other on providing the information and moral capacity for reflection, might support meaningful human control.

Our view on Reflective Hybrid Intelligent (RHI) systems goes beyond task-oriented human-agent teaming frameworks. We take a step further by focusing on moral values which require a deep reflection. Human and AI agents should together formulate the performance criteria for the RHI system. For this, we make an explicit connection to the method of WRE, in which *Moral Principles*, *Moral Judgments* and *Background Theories* are considered. In this process, one can go back and forth and alter the situation, judgements, or principles, to find a situation that expresses acceptable moral agreement.

The three elements of the WRE method form the Knowledge base for the hybrid reflection approach. Agents and humans must form a shared mental model of the three elements of WRE. However, forming such a shared mental model can be challenging. From the humans' perspective, it can be challenging to understand the underlying decision-making processes of the agents (especially when "black-box models" are used) and the way they process information is very differ from humans. From the agents' side, operationalizing complex concepts such as values and ethical principles in a meaningful and unambiguous manner is difficult. To address these challenges, explicit criteria must be defined on how the RHI system can be monitored with respect to these criteria and who is responsible for monitoring which aspects. For example, Moral Principles can be defined as deontic rules or indicative metrics, Moral Judgments can be represented by a dataset of existing morally acceptable decisions, and Background Theories can represent the limits of how agents and humans pro cess information and can (or cannot) make conscious moral decisions. Humans and agents should have shared understanding of when and how to report their findings. Further, they have an obligation to check the indicators (related to the Moral Principles, Moral Judgments and Background Theories) on a regular basis. In case of aberrations, they must have a clear procedure for initiating a contestation, and starting the reflection process as a co-activity. This is where the tracking condition for MHC (Santoni de Sio and van den Hoven, 2018) comes into play. While humans and agents may have different mental models at times, MHC can still be supported if they work together in this manner.

The MAPE-K architecture dissects the feedback loop into four parts that share the knowledge base. The Monitor step provides the necessary information to trigger, support or evaluate a reflective process. This step can be performed either by the agent or by the human. Agents can collect, aggregate and filter information as well as report metrics that relate to a moral value. For example, agents can calculate different metrics of fairness and report it to the human. Humans can provide additional information to the system, both considering quantitative or qualitative indicators. Further, in this step, both human and agent can observe an anomaly or detect an aberration from the predicted behaviour impacting a given value. The aim is to alert the other of this fact, which is tightly coupled to the explanation capabilities of either to form a shared mental model of what the contestation is about.

The next step is Analyze, which correlates and aggregates the information from the monitoring step into more complex context-dependent investigations. Critical to this step are bi-directional explanations between the agent and the human; both the human and the agent need to be able to explain issues to each other. Furthermore, the effectiveness of explanations needs to be checked, with questions which might lead to further explanations, or even to explanations of the questions. As a result, explanations in HRI systems would naturally take the form of conversations between the AI and the human. Human and AI have a joint responsibility to maintain a shared mental model of the HRI system. Of course, the AI only needs to form a shared mental model of the human's view of the HRI system and the world in as far as that pertains to the HRI system's responsibilities.

A reflection can result in a contestation that after a joint discussion leads to the insight that some behaviour of the system needs to be changed. This is where the Plan step comes into the picture. In such a case, the human and AI should engage in a discussion of what needs to change and how. Note that either can suggest that expertise is needed that is not currently part of the HRI system. The human has the final responsibility of deciding who (or what form of AI) to add to the HRI system. Following the tracing condition of MHC (Santoni de Sio and van den Hoven, 2018), this changes and the reasoning leading to any decision should be logged (including the decision that no additional expertise is needed, when this is the case). Using both human and AI expertise, decisions are made of what to change and who is responsible for that change. Note that the final agreement regarding any of such decisions has to come from the human, supported by the agent. Part of the decisions on changing the system has to be on what criteria will be used to monitor the system after the changes have been carried out and how the changing process (act) of the system will be monitored.

The Execute step provides the mechanisms to execute the plan, considering the dynamical aspects of the socio-technical environment. Given the complexity of the HRI system and the part of society it affects, also the change of the system might not be straightforward and has to be monitored, contested and reflected upon if need be, and changed (changing the change plan).

All of the above processes of the MAPE-K are part of the co-activity of human and AI in the HRI system. This means that a co-activity analysis should be performed for all of these processes. Further, this has to be done whenever a change is in order. Interesting point of research is to what extent each of the above can be implicitly implemented in the AI part of the HRI system.

## 11.4 Working example: negotiation and fairness

Consider a human *H* and a negotiation support agent *A* together negotiating with another party *P*. For this example, it doesn't matter whether the other party is a human or an AI support agent negotiating on behalf of a human. Examples of such support agents are available, e.g., in the GeniusWeb platform (Hindriks et al., 2009; TU Delft, n.d.-a), the NegMAS platform (Mohammad, Nakadai, & Greenwald, 2019; Mohammad, n.d.) and the Pocket Negotiator (Hindriks & Jonker, 2008; Jonker et al., 2016; TU Delft, n.d.-b).

Human *H* and support agent *A* together form a RHI system. Let us consider that the human *H* desires to negotiate fairly, which intuitively means that all parties should get a "good" negotiation outcome. However, what "negotiating fairly" means is a difficult question. Thus, the RHI system should reflect on the value of **fairness**. To consider the intricacies of fairness (Saxena et al., 2019), which form the core of our case study, we look into it from the multiple levels of consideration, as defined in the KNOWLEDGE base (Figure 11.2).

- **Moral Principles**. Providing a general definition of fairness is not trivial. The meaning of fairness can depend on context, cultural, and other social factors; individual differences can lead to very different conceptions. Fairness can be considered an "essentially contested concept" (Gallie, 1955), i.e., a concept that the proper use of which inevitably involves disputes about their proper uses on the part of their users. In negotiation, fairness come in different forms. Albin (1993) identifies four types of fairness on negotiation: structural fairness, process fairness, procedural fairness and outcome fairness. In this example, we focus on outcome fairness, more specifically on three possible principles that define fairness: **equity**, **equality** and **need**. The equity principle of fairness holds that the outcome should be proportional to the investments (financial or not). The equality principles holds that the outcome should be comparable to all parties, regardless of their investments or need. The need principle, also known as redistributive justice, stipulates that the outcomes should be proportional to the need of each party—the party that has the greater need gets a larger share of the pie.
- **Moral Judgments**. Fair outcomes are not necessarily the most efficient or even logical and cannot always be easily traced back to a single principle. In negotiation, fairness can be considered a social construct that is built during negotiation between both parties – a shared understanding needs to be built. In other words, fairness can be seen as an element of acceptability (Doorn & Taebi, 2018). Hence, to reflect about fairness in negotiation, we need an understanding of what the negotiating parties (*H* and *P*) deem as fair in a given context. However, as preference profiles are subjective and private, negotiating parties typically do not want to share their preference profile with other parties. Further, sometimes what they choose to disclose might not be fully in line with their preferences. This strategic behaviour might be caused by a fear of being exploited, or from with the intent to gain more than their fair share (Murukannaiah & Singh, 2020).
- **Background Theories.** Multiple normative and psychological theories can influence the establishment of a shared definition of fairness during a negotiation process. In this example, we consider two areas: the hyperbolic discounting and power relations. *Hyperbolic discounting*: In negotiations, it may not be the fairness in each bid that is important, but the fairness of the outcome. The process of negotiation can be seen as a way to find outcomes that are acceptable (w.r.t. to fairness) to all negotiators. Thus, both negotiating parties contribute to fairness by not accepting bids that are unacceptable to them. This complicates matters in that it might be counterproductive to complain that the early bids are not fair. The timing has to be close to the end of the negotiation. However, this might be in disagreement with the time-inconsistent model of hyperbolic discounting or "present bias", according to which people tend to choose immediate smaller rewards rather than later larger rewards. **Power relations**: In a negotiation, social circumstances matter. Then, what is fairness in the face of one party having more bargaining power than the other? Bargaining power is the relative ability of parties to influence each other in an argumentative situation to get what they want (Bacharach & Lawler, 1981).

*11.4.1 Reflective Pocket Negotiator*

We discuss a hypothetical RHI negotiation support system, called Reflective Pocket Negotiator (RPN) that pays attention to the intricacies of fairness we discussed above. RPN is an extension of the Pocket Negotiator (PN) (Jonker et al., 2017), which supports preference profiling of user and opponent, and assists in bidding and acceptance decisions. Figure 11.3 shows the bidding support interface of the current PN.

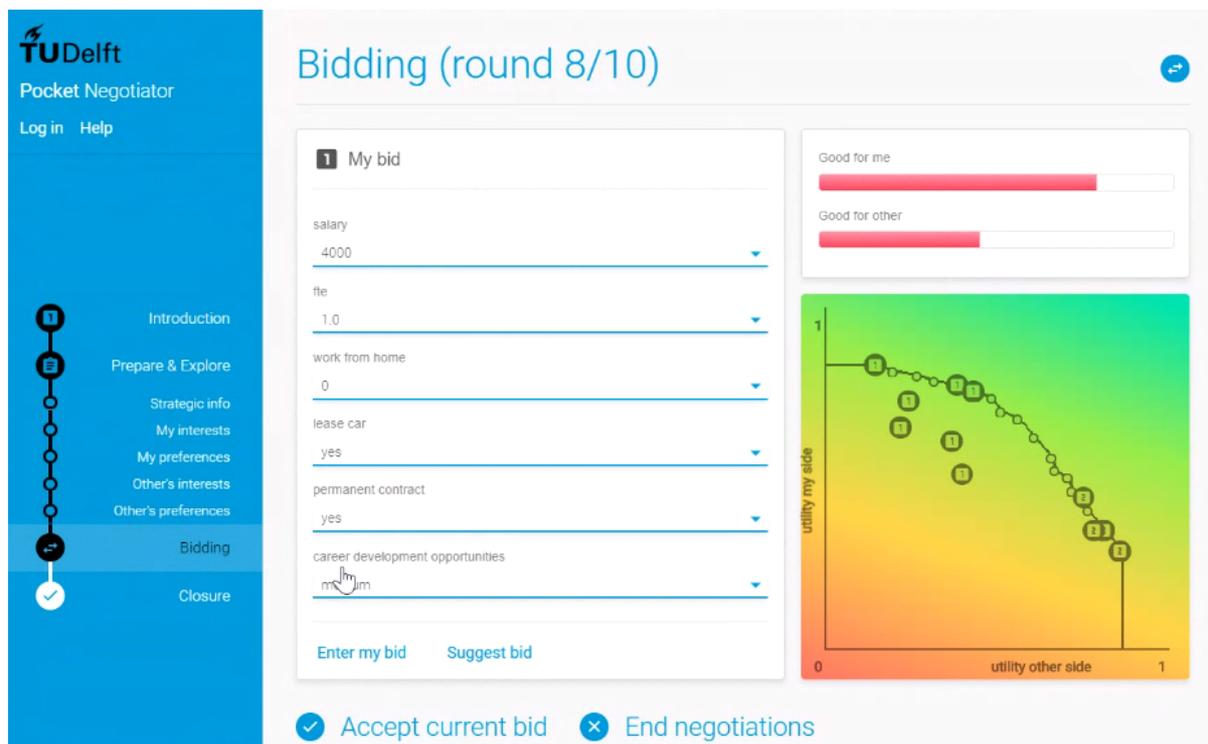

*Figure 11.3* Bidding interface of the existing current Pocket Negotiator (PN) (TU Delft, n.d.-b)

A key goal of the hypothetical RPN system is to reflect on fairness by the complimentary expertise of humans and AI components. We illustrate this by presenting a co-activity analysis (Johnson et al., 2014a) on human user ($H$) and AI ($A$) negotiating with another party ($P$) with respect to reflecting on fairness as shown in Table 11.1.

We include $P$ in the co-activity analysis of the pre-negotiation phase. Since $P$ clearly has a stake in the negotiation, it co-determines the negotiation outcome. Given that $P$ is not on the "team" of $H$ and $A$, the aspects of the co-activity analysis during negotiation are left blank for $P$.

In RPN, negotiating fairly is a joint responsibility of the user *H*, the intelligent agent *A* and, to some extent, the opponent *P*. Thus, the reflection on fairness in RPN starts by establishing a shared mental model (Jonker, Riemsdijk, & Vermeulen, 2011) of fairness among *H* and *A*, and to some extent with *P* in the pre-negotiation phase. Note that *H* cannot force *P* to fully share their opinion about fairness.

***Table 11.1*** *Co-activity analysis for negotiating fairly, where H is a negotiating human (primary task performer), A is H's negotiation support agent, and P is the party (human or agent) with whom H is negotiating. Legend for the cell colours:* I can do it all, I can do it all but my reliability is less than 100%, I can contribute but need assistance, I cannot do it.

| Task | Subtask | Interdependencies | | |
|---|---|---|---|---|
| | | *H* | *A* | *P* |
| Pre-negotiation | Model the negotiation domain | | | |
| | Select a negotiation protocol | | | |
| | Define principle for fairness | | | |
| **Monitor** fairness | Record bids | | | |
| | Compute fairness metrics | | | |
| **Assess** fairness | Analyze fairness metrics | | | |
| **Plan** fair negotiation | Elicit *H*'s preferences | | | |
| | Estimate *P*'s preferences | | | |
| | Adopt a fair bidding strategy | | | |
| | Adopt a fair acceptance strategy | | | |
| **Execute** fair negotiation | Bid, accept or reject | | | |

As Table 11.1 shows, the activities involved in the negotiation align with the MAPE (Monitor, Assess, Plan and Execute) loop. Further, most of the activities involve interdependencies between the human *H* and the agent *A*.

The agent *A* requires formal criteria of fairness in order to monitor fairness during the negotiation. Depending on the abilities of the user *H*, either they select those criteria from the fairness criteria presented by *A*, or they formally define these criteria themselves. Although some of the available bidding strategies available in the literature try to optimize the utility for all negotiating parties, that does not imply that the outcome is fair. None of the currently available bidding strategies in the original PN explicitly provide a definition of fairness that can be communicated to humans. Therefore, these definitions should be added to the RPN, along with the required explanation power.

Being an extension of PN, RPN will provide logging of the bids exchanged and the current estimated preference profiles (feature already existing in the current version). These can be used in RPN to mathematically assess the fairness with respect to the chosen criteria indicative of fairness. The complicating factors as mentioned above are that these definitions do not account for the social situation, the subjectivity of preference profiles, and the inaccuracies in the estimation of the opponent's preference profile. Therefore, only monitoring the fairness on the basis of a mathematical criteria indicative of fairness is not enough; the humans will have to actively think and discuss the concept of fairness themselves.

During the negotiation, humans may have problems maintaining an overview of the preference profiles and how to assess each bid in terms of these profiles. In RPN, as it extends PN, agent *A* supports the human *H* in this by providing a picture that plots each bid in a two-dimensional graph based on the estimated utilities of both negotiators (see right-hand side of Figure 11.3). Further, it provides red bars that provide similar information in a different format. In terms of utility, the red bars show that the current offer of the user (*H*) would be rather better for themselves than for their opponent (*P*).

RPN's agent *A* is not assumed to have the functionality to monitor the conversation of the user *H* and their opponent *P*. Thus, any information exchanged by them has to be interpreted by *H*. For example, *H* might learn that the preference profile of *P* is different than estimated so far, or that *P* is fine with not getting the best of the deal this time and is satisfied by a future stake on *H*, to restore the balance.

In RPN, the user *H*, with some help of agent *A*, can make an assessment of the fairness of each bid and of the negotiation outcome. The timing of when to check for fairness is an interesting challenge. Who knows when the negotiation is about to end? Given the current state of the art in negotiation support systems, it is the human who is best capable of that assessment, as none of the currently available systems have functionality to support this. Cultural differences do play a role, see e.g., (Graham, 1985; Hofstede, Jonker, & Verwaart, 2012; Salmon et al., 2016; Kong & Yao, 2019). Training a machine learning algorithm for this might seem an obvious way to go. However, gathering data for this purpose will have to be across many negotiation situations with many different humans, social situations and cultures and negotiation domains, which will lead to rather non-specific results. To get specific results, the agent could learn over many negotiations in the same domain, in similar cultural settings. If that is successful, one might extend learning over different contexts (getting a grip on the cultural dimension), and over negotiators from different cultures for negotiations in the same domain.

*11.4.2 Case study: hybrid reflection with MAPE-K in RPN*

Consider that the user *H* initiates activities of the hypothetical RPN by indicating that a negotiation is at hand. As *H* is a new user, Agent *A* explains that to better support *H*, they need to establish first what the negotiation is about (domain of negotiation) and what each of the negotiating parties would like to get out of the negotiation (preference profiles). The details of how that could be done are beyond the scope of this paper, but do fit within the existing PN functionality. Basically, these activities are the essence with which *A* and *H* establish a shared mental model of the negotiation domain and the preference profiles of *H* and what *H* estimates about *P*'s preferences.

We skip the part of the scenario in which *A* explains the reflective functionality it aims to execute in collaboration with *H*. We pick it up at the point where agent *A* has asked *H* whether *H* has some specific wishes about the outcome. In an interface that *A* provides, *H* is asked to indicate which properties *H* would like to reflect on together with *A*. *H* can select some existing properties, such as fairness, but it can also indicate that a new property has to be formulated. In our case study, *H* selects fairness. In the hypothetical RPN, agent *A* would continue by explaining that during the negotiation: *A* will monitor the offers made by *H* and *P*, and analyze how it relates to the elements of the knowledge base outlined in the outset of this section. In case of doubt, *A* will raise the issue with *H* to enable a joint reflection on *P*'s preferences.

*A* presents another interface, now with some pre-defined conceptions of fairness principles, namely: equity, equality, and needs. For example, the explanation about "needs fairness" emphasizes the social situation and needs of all parties. To explain this better, the agent *A* presents an example of distribution of goods (say food and clothing), saying that if one party has a whole family to feed and is poor, then it would be fair to give most of the goods to that party. In contrast, if *H* considered the equity principle, *A* would provide a business case where it is fair to pay the market value and not more or less than that.

The agent *A* continues by explaining that, whatever choice *H* makes, *A* will help *H* in translating the chosen fairness principle into a preference profile over the negotiable issues. This approach, which we call a value-based preference profile, is already partially deployed in PN, and is an extension of the interest-based negotiation approaches put forward by Fisher and Ury (1981), and refined and commented upon by many researchers. For example, Mnookin et al. (2000)

brought forward that negotiation should be about creating value for all parties; Wolski (2012) explained how the skill of the negotiator can lead to an unethical difference in outcome, skewed to the side of the skilled negotiator; Rahwan et al. (2003) provided a mathematical framework, not unlike to what is available in PN (Jonker et al., 2016).

As fairness is an essentially contested concept (Gallie, 1955), its meaning can be disputed by other parties, in our case by the opponent $P$. Consider a scenario in which $H$ wants to adopt the "needs principle" to monitor and operationalize fairer bids. In this case, $H$ assumes to be in a position of advantage towards $P$ (i.e., $H$ is less impacted by the outcomes of the negotiation than $P$); thus, $H$ aims to provide a better deal for $P$, while still trying to achieve its goals.

To support $H$ in deciding on which bid to select, the RPN presents two additional lines in the bidding space as indicated in Figure 11.4(a). In this figure, we call the dotted line the Line of Equal Opportunity, inspired by Kalai (1977), which corresponds to the equality principle under the assumption of normalized utilities (see Section 11.4.3). The user's attention is directed to the blue dots, which refer to potential outcomes of the negotiation which are closest to that line and for which both parties would assign almost equal utility. The point where this Line of Equal Opportunity crosses the Pareto Optimal Frontier, is called the Egalitarian point. The figure also displays a dashed line with green points close to it. The dashed line is what we here introduce as the Line of Balanced Needs, i.e., the green points are those potential outcomes in which the benefit for the negotiators is balanced with respect to their needs.

From a negotiation theoretical point of view, one might think that this corresponds to party $P$ having a higher reservation than $H$. However, $P$ might be in such need that any deal would be better than no deal at all, which would mean that $P$ would have a very low reservation value. Similarly, $H$, being "rich" anyway, might have a high reservation value. So, fairness from a "needs principle" cannot be defined in terms of the concept of a reservation value.

In this example, if $H$ were negotiating with a much richer person, it would feel that the Line of Balanced Needs would be on the other side of the Line of Equal Opportunity. Then, what the agent $A$ offers to $H$ is to rotate the outcome space in such a way that the Line of Balanced Needs coincides with the diagonal. $A$ explains that by doing that, $H$ can make use of bidding strategies that aim for social welfare outcome even though these strategies were originally created for social welfare in terms the principle of equal utility. $H$ accepts this proposal and the bidding starts.

Now, assume that, by monitoring of the bids of $P$, agent $A$ believes that the bids proposed by $P$ seem to steer towards an outcome of similar utility values for both parties, i.e., according to the equality principle. Agent $A$ communicates this reflection to $H$ and provides evidence of said behaviour of $P$ to $H$, namely that $P$ rejected a bid made by $H$ that clearly favoured $P$ in terms of utility. After reading this feedback $H$ reflects that $P$ does not seem happy with how the negotiation is going and decides to discuss the matter with $P$. In that discussion, $P$ indicates that accepting a bid that is not fair in terms of equality is unacceptable. $H$ realizes that $P$'s pride is hurt, and repairs the negotiation by enlarging the negotiation space by adding a side job that $P$ could do and that would really help $H$. By doing that, the balance in needs now corresponds to a balance in utility. In other words, in the new bidding space the Line of Equal Opportunity, now coincides with the Line of Balanced Needs. Note that the green dots in the original bid space shift towards the Line of Equal Opportunity, when adding in those bids that the side job will be done by $P$. To further enhance the RPN, it may be useful to have the system explicitly state whether or not an offer is fair according to a given principle. Human negotiators could then provide feedback by agreeing or disagreeing with this claim, and in cases of disagreement, be able to explain to the system why they believe an offer is unfair. This can be a challenging task, especially when opinions on fairness are not fully aligned with well-known principles such as equity, equality and need. In such cases, a co-active learning approach may be adopted to elicit and learn the human negotiators' values. This approach, although challenging, could allow for a deeper understanding of the negotiators' beliefs about fairness and enable the system to adapt to their values.

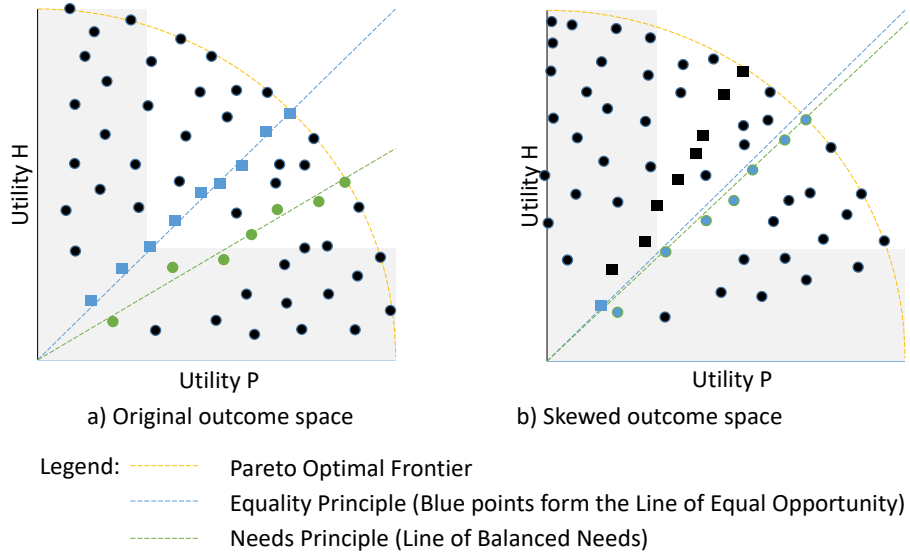

*Figure 11.4* An agent A explaining the rotation of the bidding space to a human H during the reflection process (a) from the Equality Principle to the Needs Principle, and (b) how the Needs Principle can coincide with the Equality Principle by enlarging the space.

*11.4.3 Theoretical side note*

The Egalitarian Point or the Rawls point (after the philosophical theory of Rawls (2003)) also known as the Kalai point (not to be confused with the Kalai-Smorodinsky point), is a lesser-known bargaining solution. It was proposed by Kalai (1977) two year after the Kalai-Smorodinsky Solution. Intuitively, the Egalitarian point (*EP*) tries to maximize the utility of the party with the lowest utility. Under the assumption that all utility functions are normalized, (*EP*) can be found by choosing the point $a$ on the Pareto Optimal Frontier $F$ with the highest minimum utility $u_p$ for all negotiating parties $p \in P$:

$$EP = \operatorname*{argmax}_{a \in F} \operatorname*{argmin}_{p \in P} u_p(a)$$

For more details and examples on the proportional solution that underlies this point, we refer to Kalai (1977). In a normalized utility space, the Origin is the point where all negotiators would receive the minimum utility of 0. Similarly, we define the Utopian point, as the point where all negotiators would receive the maximum utility of 1. Let $D$ be the diagonal of the normalized utility space *US*; the straight line passing through the origin and the Utopian point, which in a bilateral negotiation would be the line $x = y$. We define the Line of Equal Opportunity (LEO) in the normalized utility space to be the line of points that correspond to bids $b$ from the bid space $B$ for which the utilities of the negotiating parties are closest under the Euclidean distance (denoted by $d$) to the diagonal $D$:

$$LEO = \{b \in B \mid \exists_a \in D : b = \operatorname*{argmin}_{b' \in B} d(u(b'), a)\}$$

Note that for a bid space in which all issues have a continuous range of values, *LEO* corresponds to *D*.

Finally, *H* adding a side job to each bid to help *P* yields a new bid space *B'*. In this new bid space, the Line of Balanced Needs is closer to (and ideally coinciding with) the Line of Equal Opportunity compared to the original space *B*. However, defining the new bid space precisely requires characterizing the utility of the side job for both parties, which is out of scope for this chapter.

**11.5 Conclusion**

In this chapter, we introduced the concept of Reflective Hybrid Intelligence (RHI), in which a human and an AI agent mutually reflect on a desired value of an intelligent system. Our conception of RHI founded on the philosophical method of Wide Reflective Equilibrium (WRE), which includes the interplay among moral principle, moral judgements and background theories. Further, we integrate WRE into a computational framework via the MAPE-K loop in order to facilitate AI agents to participate in reflection. We demonstrate our ideas via a use case on Reflective Pocket Negotiator (RPN), an intelligent system for negotiation support. In particular, we explore how a human and agent reflect on the value of fairness, e.g., considering the principles of equity, equality and need, in RPN. Our contribution provides a solid theoretical framework for developing RPN and similar decision support systems such that humans remain in meaningful control of AI in intelligent systems.

**Notes**

[1] Tracking condition: to be under meaningful human control, an AI system should be responsive to the human moral reasons relevant in the circumstances.

[2] Tracing condition: in order for an AI system to be under meaningful human control, its actions/states should be traceable to a proper moral understanding on the part of one or more relevant human persons who design or interact with the system.